\pdfoutput=1

\documentclass[11pt]{article}

\usepackage[review]{acl}

\usepackage{multirow}
\usepackage{times}
\usepackage{latexsym}
\usepackage{color}
\usepackage[T1]{fontenc}
\usepackage{subfigure}
\usepackage{graphicx}
\usepackage{amsmath}
\usepackage{amsfonts}
\usepackage{enumitem}
\usepackage{algorithm}
\usepackage{algorithmic}

\usepackage[utf8]{inputenc}

\usepackage{microtype}
\usepackage{booktabs}
\title{Wav2SQL: Direct Generalizable Speech-To-SQL Parsing}

\author{
  Huadai Liu$^1$\thanks{Equal Contribution.}  \And Rongjie Huang$^1$\footnotemark[1] \And \qquad Jinzheng He \And \qquad Gang Sun \And \qquad Ran Shen \And \qquad Xize Cheng \And Zhou Zhao$^1$\thanks{Corresponding Author.}\\ $ $\\
$^1$Zhejiang University   \qquad $^2$Zhejiang University 
}

\begin{document}
\maketitle
\begin{abstract}
    Speech-to-SQL (S2SQL) aims to convert spoken questions into SQL queries given relational databases, which has been traditionally implemented in a cascaded manner while facing the following challenges: 1) model training is faced with the major issue of data scarcity, where limited parallel data is available; and 2) the systems should be robust enough to handle diverse out-of-domain speech samples that differ from the source data. In this work, we propose the first direct speech-to-SQL parsing model Wav2SQL which avoids error compounding across cascaded systems. Specifically, 1) to accelerate speech-driven SQL parsing research in the community, we release a large-scale and multi-speaker dataset MASpider; 2) leveraging the recent progress in the large-scale pre-training, we show that it alleviates the data scarcity issue and allow for direct speech-to-SQL parsing; and 3) we include the speech re-programming and gradient reversal classifier techniques to reduce acoustic variance and learned style-agnostic representation, improving generalization to unseen out-of-domain custom data. Experimental results demonstrate that Wav2SQL avoids error compounding and achieves state-of-the-art results by up to 2.5\% accuracy improvement over the baseline.
\end{abstract}

\section{Introduction}

Speech-to-SQL parsing (S2SQL) aims to generate the SQL query from a spoken question based on relational databases, which is highly beneficial for breaking down barriers among those who are not well proficient in SQL queries. Conventional S2SQL systems~\citep{kumar2013system,song2022speech} are often composed of a cascade of two components: automatic speech recognition (ASR)~\cite{yu2016automatic,schneider2019wav2vec,hsu2021hubert} and text-to-SQL parsing~\cite{bogin2019representing,bogin2019global,chen2020measuring,guo2019towards}. Compared to cascaded systems, work on direct S2ST is very limited, with the potential benefits of 1) working on languages without written form~\cite{campbell2008ethnologue}, where an estimated half of the 7,000 languages in the world actually do not have written forms; 2) avoiding error compounding across sub-systems~\cite{nakamura2006atr,jia2019direct}.

The recent development of direct S2SQL parsing still faces several challenges: 1) despite the benefits of direct approaches, model training is faced with the major issue of data scarcity. Human-labeled speech data is expensive to create, there are very few data resources providing parallel speech, and the data amount is quite limited, 2) increasing demand for SQL parsing from personalized speech challenges models especially in unseen scenarios. When the distributions of custom voice (speaker and accent) differ from training data, the system performance deteriorates due to distribution gaps, and 3) the modality gap between the spoken question and text schema hinders the ability of the question schema, making it difficult to align question speech to the intended tables.

To accelerate S2SQL research, we assemble an open-source, multi-speaker, and multi-accent S2SQL corpus MASpider. To the best of our knowledge, MASpider is the first open-source speech-to-SQL parsing dataset. We have attached part of MASpider to the
supplementary materials, and we will release the entire dataset after the paper publication. To overcome the aforementioned challenges in this paper, we propose Wav2SQL for direct speech-to-SQL parsing, which is generalizable to unseen acoustic conditions (speaker and accent) in custom data. To be more specific, 1) leveraging self-supervised learning (SSL)~\cite{baevski2020wav2vec,hsu2021hubert}, it alleviates the data scarcity issue and benefits S2SQL model training, 2) we introduce speech re-programming and gradient reverse technique to effectively eliminate the style attributes in representation, which promote the model generalization to unseen speakers and accents in custom data.

Experimental results on the MASpider dataset demonstrate that our Wav2SQL model surpasses the cascaded system in the exact match accuracy and achieves competitive performance with our model trained on the TTS dataset. The main contributions of This work are summarized as follows:
\begin{itemize}
    \item We introduce the first cross-domain speech-to-SQL parsing benchmark dataset MASpider\footnote{Audio samples are available at \url{https://Wav2SQL.github.io/}}. 
    \item Leveraging self-supervised learning, we propose the first direct speech-to-SQL parsing and show that the large-scale pre-training alleviates the data scarcity issue.
    \item Through introducing speech reprogramming and gradient reversal technique, we effectively eliminate the style attributes in speech representation and predict the style-agnostic variation, which significantly improves the model generalization to unseen speakers and accents in custom data.
    \item Experimental results on the MASpider dataset demonstrate that our model outperforms the cascaded systems and achieves state-of-the-art performances.
\end{itemize}

\section{Related Works}

\subsection{Text-to-SQL Parsing}

Semantic parsing of natural language to SQL query recently surged in popularity because of the release of two cross-domain datasets-WikiSQL~\cite{zhong2017seq2sql} and Spider~\cite{yu2018spider}. IRNet~\cite{guo2019towards} encodes the question and schema via bi-LSTM and proposes the string match strategy for schema linking. RATSQL~\cite{wang2019rat} presents a unified framework with a relation-aware transformer(RAT) to encode relational databases and NL questions. SADGA~\cite{cai2021sadga} adopts the graph structure to provide a unified encoding model for both the NL question and databases. In recent years, speech-to-SQL systems usually adopt cascaded automatic speech recognition with text-based SQL parsing. However, the error propagation hurts model performance, not to mention that numerous languages do not have written forms. In this work, we present the first direct speech-to-SQL parsing model without using text, which demonstrates the generalization to different accents and speakers.

\subsection{Self-Supervised Learning in Speech}

Self-supervised speech representation learning encodes the speech feature into context representations. TERA~\cite{liu2021tera} learns speech representation by reconstructing acoustic frames from their altered counterparts. Vq-wav2vec~\cite{baevski2019vq} learns discrete representations via a context prediction task using contrastive loss. Similarly, wav2vec 2.0~\cite{baevski2020wav2vec} is an end-to-end version of vq-wav2vec, while HuBERT~\cite{hsu2021hubert} predicts masked frames pre-quantized using k-means. In this work, we leverage the recent success of self-supervised learning in speech and show that large-scale pre-training alleviates the data scarcity issue and benefits model training.

\subsection{Domain Generalization}

Domain generalization aims to learn domain-invariant knowledge which can be generalized to target distribution, which attracts attention from researchers~\cite{zhou2020deep,shi2021gradient,tian2022neuron,huang2022generspeech,huang2021multi}. ~\cite{li2018deep} propose a conditional invariant adversarial network to learn class-wise adversarial networks and ~\cite{zhao2020domain} learns domain-invariant features by introducing additional entropy regularization to minimize the KL divergence between the conditional distributions of different source domains. For spoken language understanding, unseen speakers and accents in custom data significantly hurt model performance due to the distribution gaps. In this work, we introduce speech reprogramming and gradient reverse to disentangle semantically irrelevant information, leading to the significant promotion of model generalization to custom scenarios.

\begin{figure*}[ht]
    \centering
    \small
    \vspace{-4mm}
    \hspace{-5mm}
    \subfigure[Train/Test Split]{
        \label{fig:split}
        \includegraphics[scale=0.532]{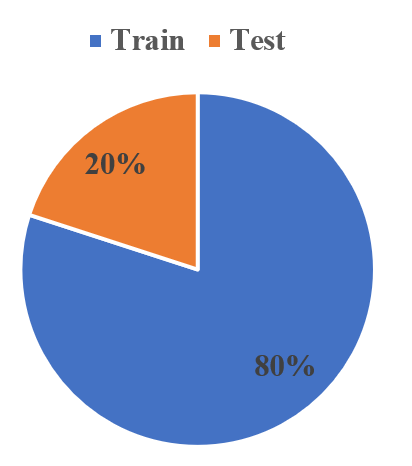}
    }
    \subfigure[Gender Statistics]{
        \label{fig:gender}
        \includegraphics[scale=0.532]{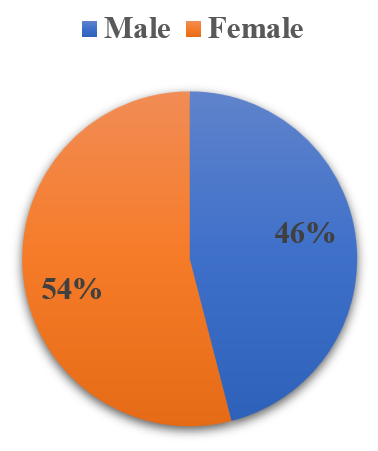}
    }
    \subfigure[Visualization of different country distributions]{
        \label{fig:country}
        \includegraphics[scale=0.5]{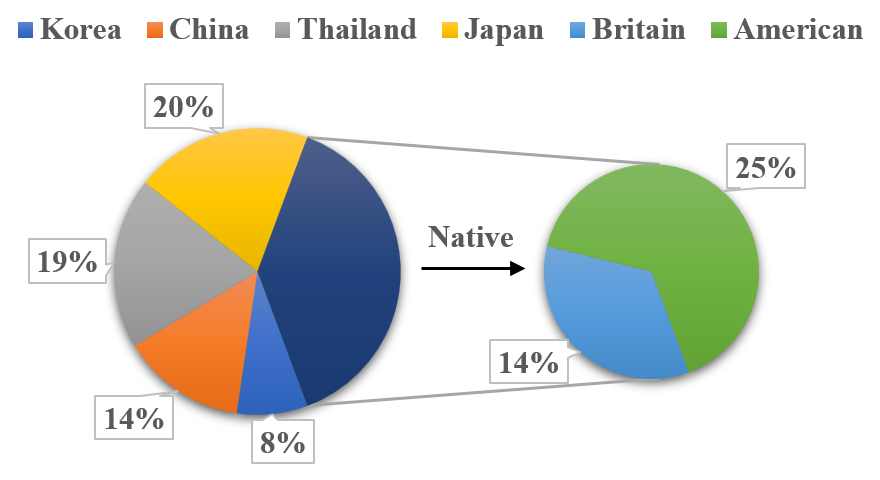}
    }
    \hspace{-5mm}
    \caption{The statistics for MASpider. }
    \vspace{-4mm}
    \label{fig:stat}
\end{figure*}

\section{Dataset Construction}
We build MASpider upon the Spider~\cite{yu2018spider} dataset which provides textual inputs, databases, and SQLs parse trees. It consists of 9693 spoken utterances recorded in a quiet environment, including 11 speakers from six different countries. MASpider consists of 15 hours of speech samples recorded in a professional recording studio, including 8.1 hours from 6 females and 6.9 hours from 5 males apart from the person-of-interest (POI). Figures~\ref{fig:stat} summarize the distribution of dataset split, gender, and country. The major features of MASpider include:

\begin{itemize}
    \item Open source. A lack of data could hinder the construction of speech-to-SQL systems, so we release our corpus to accelerate research in the community.
    \item Multi-accent and multi-speaker. Since the distribution of custom voice could be different from training data, we construct a multi-speaker and multi-accent dataset to improve model generalization.
    \item High quality. High-quality audios without noise or background sound are essential for speech-to-SQL parsing. A professional and quiet environment ensures high-quality utterances in MASpider.
\end{itemize}

\subsection{Data Collection and Verification}

\textbf{Collection Procedure}
For all 9693 utterance-SQL pairs in MASpider, we ensure that each speaker is assigned no more than 1500 sentences to avoid excessive data distribution bias. Next, we collect the audio sample of the given text utterances in a professional recording studio. Finally, the spoken utterances are saved in wav format, sampled at 16kHz, and quantized by 16 bits.

\textbf{Data Labeling}
For further study, we tag additional statistics such as the native language and age of the speakers, and the year of their English study. Following this, the dataset is split into 12-hour spoken questions for training, additional 3-hour utterances with unseen accents, speakers, and databases for testing, which enable the evaluation of model generalization to custom data. Figure~\ref{fig:split} illustrates the distribution of the training and test sets on MASpider.

\textbf{Data Verification}
Firstly, we check that the accent in the recording matches the speaker's country. Then, We listen to every recording to check for mispronounced errors and re-record the recording with more than two mispronunciations. Finally, we run the preliminary qualified recordings through an ASR system to control the recorded audio quality. In our case, we used the fine-tuned wav2vec 2.0 ASR model to filter out recordings with their character error rates higher than 25$\%$. For audio with these error rates above the threshold, it is discarded and recollected again until passed.

\begin{figure*}[htbp]
    \centering
    \vspace{-8mm}
    \includegraphics[width=0.6\textwidth,trim={1.0cm 0cm 1.0cm 0cm}]{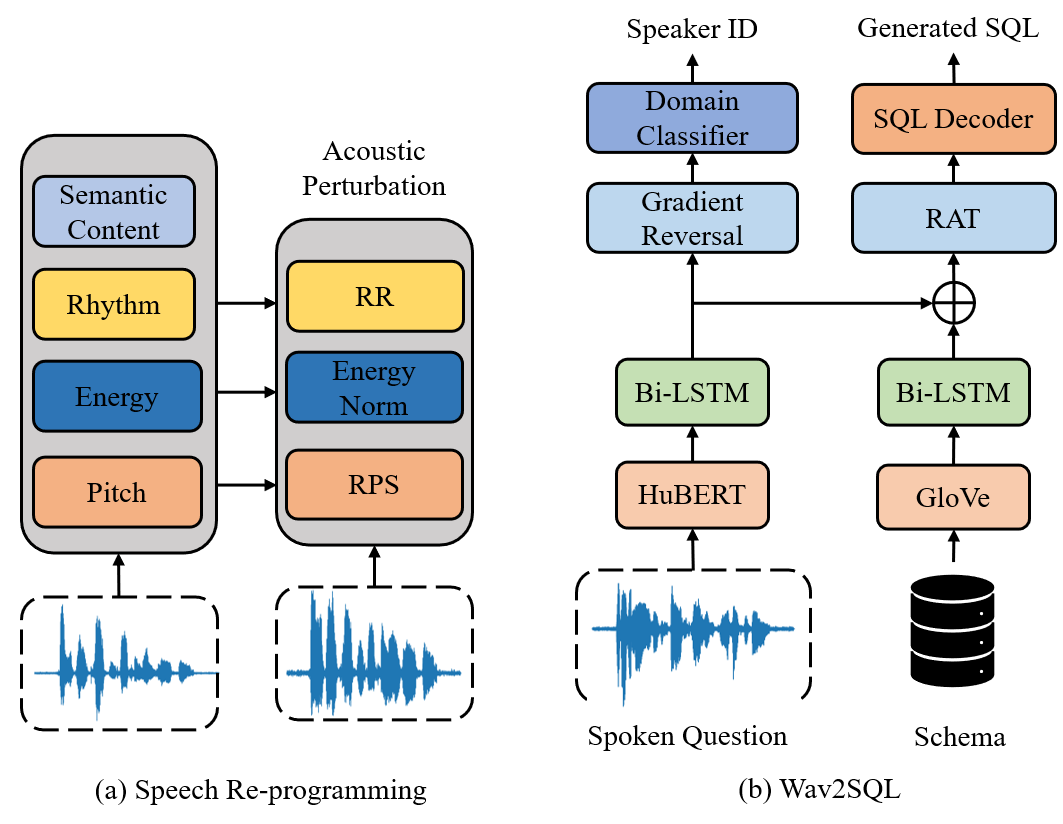}
    \vspace{-3mm}
    \caption{The information flow with dotted lines is included during training. Subfigure(a) denotes the implementation process of Speech Re-programming $RR$: random resamling; $RPS$: a chain function for random pitch shifting of raw waveform. Subfigure(b) is the overall architecture of our Wav2SQL $RAT$: relation-aware transformer.}
    \label{fig:2}
\end{figure*}

\subsection{Dataset Statistics}
After the data collection and processing procedure, we check for audio quality and conduct the statistical evaluation.

\textbf{Gender} The visualization of gender statistics is displayed as Figure~\ref{fig:gender}. As we can see, the ratio of male to female speakers is relatively average.

\textbf{Country} The recorders mainly include 4 English native speakers and 7 non-native speakers from Japan, China, Thailand, and Korea. We count the proportion of utterances recorded by these speakers and visualize it as shown in Figure~\ref{fig:country}

\textbf{Group By Difficulty}  Following the common practice~\cite{yu2018spider} to better demonstrate the model performance on different SQL queries, we group the difficulty of each spoken question into 4 levels according to the number of SQL components, selections, and conditions. Specifically, SQL queries that contain more keywords (e.g., GROUP BY, ORDER BY, INTERSECT, etc.) will be considered harder. In the end, The test set of MASpider consists of 25.5$\%$ easy, 37.9$\%$ medium, 20.9$\%$ hard, and 15.7$\%$ extra hard SQL queries. 
\section{Proposed Method}
\label{4}
\subsection{Overview}
\label{4.1}
The overall architecture has been presented in Figure~\ref{fig:2}{b}. To alleviates the data scarcity issue~\citep{huang2022singgan,huang2022fastdiff}, we leverage the large-scale self-supervised models including Hubert~\cite{hsu2021hubert} for the spoken question and language model for the textual schema to derive discriminative representation, enabling direct speech to SQL parsing. For generalizable speech to SQL parsing, we propose several techniques to promote model robustness for unseen (speaker and accent) custom data: 1) we re-program acoustic attributes and perturb the style information in speech, selectively extracting only the linguistic-related information for domain-agnostic modeling; 2) we include gradient reversal classifier to eliminate speaker information with an auxiliary gradient reversal classifier.

In the end, the tree-structure decoder produces results with an abstract syntax tree (AST) in depth-first traversal order. The training procedures are included in Section~\ref{4.5}, and more information has been attached in Appendix~\ref{sec:arch}.
\subsection{Enhanced Speech Encoder}
\subsubsection{Self-Supervised Pre-training}
\label{4.2.1}
To alleviate the data scarcity issue~\citep{huang2023make,huang2023audiogpt} and learn linguistic content from raw waveform~\citep{huang2022transpeech,huang2022prodiff,lam2021bilateral}, we leverage recent progress in large-scale self-supervised learning with Hubert~\cite{hsu2021hubert}, with a multi-layer convolution waveform encoder to generate the feature sequence followed by a Transformer~\cite{vaswani2017attention} context encoder to build the contextualized representations. 

We adopt the Hubert-Base model as speech representation, which is pre-trained on 960 hours LibriSpeech~\cite{panayotov2015librispeech}. Notably, speech representations~\citep{choi2021neural} are found to merge not only rich acoustic information but also acoustic attributes related to accents and speakers. To reduce domain-specific variations for better generalization, we investigate a novel technique in the following parts, which effectively eliminates accent and speaker information in speech representations while preserving linguistic content.

\subsubsection{Analysis: Audio Representation Quality
 Across Layers}
\begin{figure}[h]
    \centering
    \includegraphics[width=0.5\textwidth]{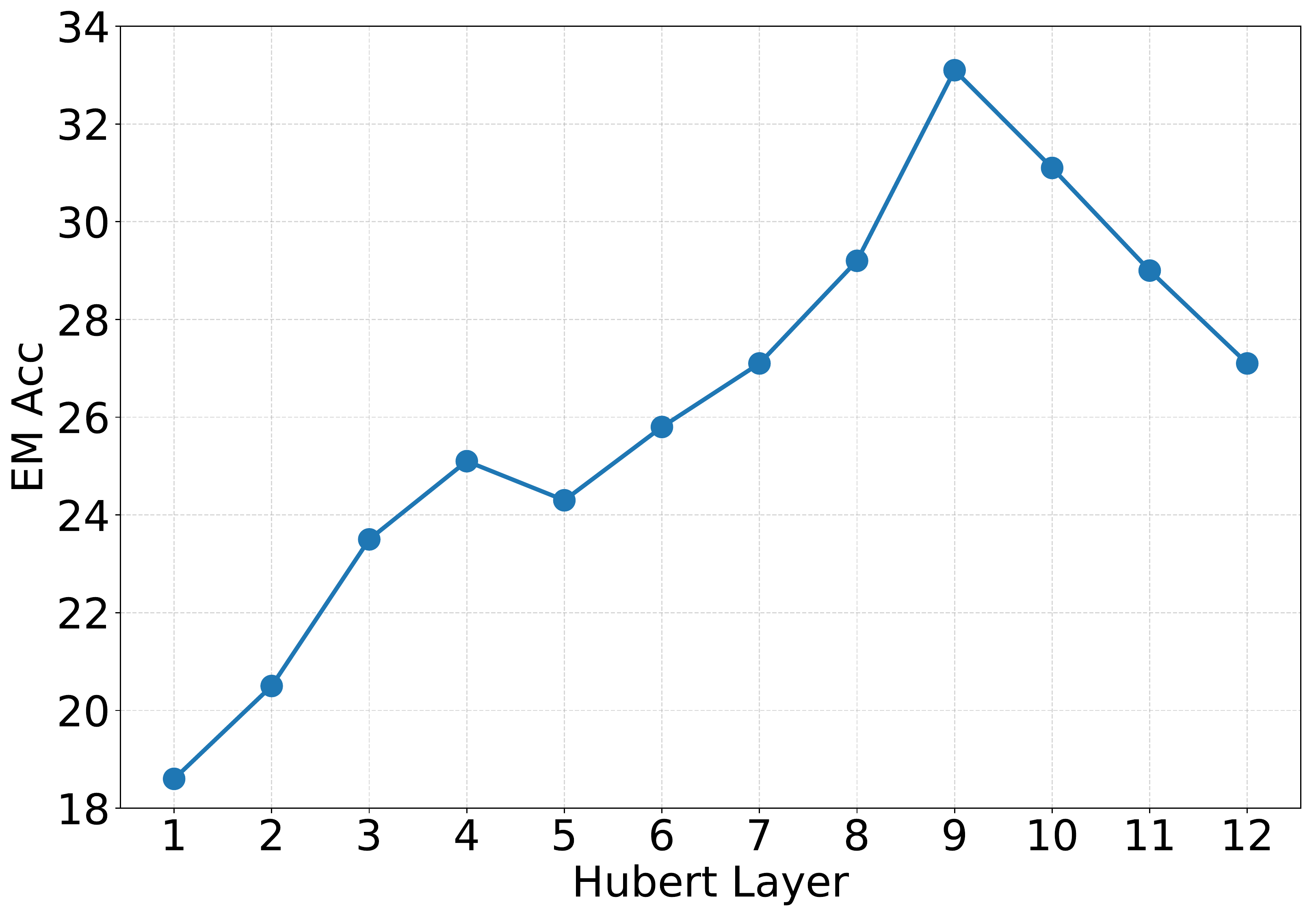}
   \caption{Speech-to-SQL generation using representations from different Hubert layers pretrained on LibriSpeech. EM Acc: Exact match accuracy. } 
    \label{fig:layer}
  \end{figure}

Before introducing our techniques for removing accent and speaker information, we first discuss the impact of the selection of different layers of Hubert on the model performance. Similar to natural language understanding, exploring the transformer layers of the BERT model shows that the underlying blocks encode syntactic information, while high-level semantic information appears in higher blocks. To make a more intuitive sense of this, we separately extract frozen representations of Hubert's 12 layers as audio features. We then input these audio features into the S2SQL model and evaluate their performance by exact match accuracy. Figure ~\ref{fig:layer} demonstrates that the first 7 layers as well as the final layer have poor performance compared to layers 8 to 11 whose accuracy is higher than 29.0 \%. Layer 9 achieves the best accuracy of 33.1 \%. 

\subsubsection{Re-program on Acoustic Condition}
\label{4.2.2}
An intuitive way~\cite{li2018domain,bui2021exploiting} to achieve better generalization is to decompose a model into the domain-agnostic and domain-specific parts via disentangled representation learning and eliminate the domain-specific variations.

In contrast, the representation derived from self-supervised models contains not only rich linguistic content but also information related to pitch and speaker, which are style-specific attributes that may decrease model generalization. As such, we conduct re-programming on speech attributes and perturb the rhythm, pitch, and energy information, which disentangle acoustic variation and selectively extract only the linguistic-related information, exhibiting better generalization to unseen custom data. As shown in Figure~\ref{fig:2}{a}, we apply bottlenecks on acoustic conditions and create re-programmed speech samples. Additional details have been attached in Appendix~\ref{sec:ap}.

\paragraph{Rhythm} Rhythm characterizes how fast the speaker utters each syllable. To perturb rhythm information, we adopt random resampling $RR$ to divide the input into segments of random lengths, and we randomly stretch or squeeze each piece along the time dimension. 

\paragraph{Pitch} Pitch is an indispensable component of intonation. First, We normalize the pitch contour to a common mean and standard deviation, removing the timbre variations in speech. Secondly, a chain function is adopted to randomly shift the pitch contour. 

\paragraph{Energy} Energy represents the magnitude of the raw waveforms and visually reflects the volume of the speech. We re-program energy attributes and create samples with different energy distributions.

\subsection{Gradient Reversal Classifier}
\label{4.3}
To eliminate the speaker identity in speech representation, we introduce a gradient reversal layer (GRL) in speaker classifier~\citep{ganin2016domain}, which regards speaker variations as a classification problem and directly maximizes the loss of the domain classifier by reversing its gradients. 
In backpropagation, GRL takes the gradient from the subsequent layer and changes its sign by multiplying with -1 before passing it to the preceding layer:
\begin{equation}
    R(x) = x,  \frac{dR}{dx} = -\textbf{I},  \label{1}
\end{equation}
where I denotes an identity matrix.

It ensures that the feature distributions between different speakers are similar (i.e. as distinguishable as possible), resulting in speaker-agnostic features. Therefore, we can further obtain audio features that preserve semantics regardless of accent and speaker, demonstrating better generalization to custom data in SQL decoding.

\subsection{SQL Decoder}
\label{4.5}
The SQL decoder follows the grammar-based architecture of ~\cite{yin2017syntactic}, which generates the SQL as an abstract syntax tree(AST) in depth-first traversal order. The generation process of SQL AST is factorized into sequential actions, which are divided into two cases: (1) \textsc{ApplyRule} which expands the last generated node according to the grammar rules or completes a leaf node, (2) \textsc{SelectColumn} and \textsc{SelectTable} represent that selects a column or table item from the schema respectively.

Firstly, the probability of generating a SQL y is defined as:
$$p(y|x) = \prod_t{p(a_t|x,a_{<t})} $$
where $x$ is the encoded memory of questions, columns, and tables, $a_t$ is the action token at time step t, and $a_{<t}$ is the sequential actions before time step t. Then in the tree-structured LSTM decoder, the hidden states at each time step t are updated as $m_t,h_t = \rm{LSTM}([a_{t-1};p_t;c_t;n_t], m_{t-1},h_{t-1})$, where $m_t$ is the cell state of time step t, $h_t$ is the hidden state, $a_{t-1}$ is the previous action embedding, $p_t$ is the parent information of the current node, $c_t$ is the context vector, and $n_t$ is the embedding of the current node type. The parent information contains the hidden state and embedding of parent action. The context vector is computed using multi-head attention upon $x$ and $h_{t-1}$. Finally, how action probabilities $p(a_t|x,a_{<t}$ are computed are explained as follows: \\
\\
For \textsc{ApplyRule} action, 
\begin{equation}
p(a_t=\rm{AR}[r]|x,a_{<t}) = softmax_R(g(h_t))
\end{equation}
where $\rm{AR}$ is the \textsc{ApplyRule} action and $g(\cdot)$ is the feed-forward network that is composed of two linear layers and a $tanh$ activation function.
\\
For \textsc{SelectTable} action,
\begin{equation}
    \gamma_j = softmax_j(\frac{(h_tW_Q)(x_jW_K+\mathbf{R}_{ij})^T)}{\sqrt{d}}),
\end{equation}
\begin{equation}
    p(a_t=\rm{ST}[i]|x,a_{<t}) = \sum_j{\gamma_j}
\end{equation}
where ST denotes \textsc{SelectTable} action. The calculation of \textsc{SelectColumn} action is similar.

\subsection{Training and Inference}
\label{4.6}
We formulate speech-to-SQL parsing as a sequence-to-tree generation problem. The input is the question (audio) and schema (text), which belong to two different modalities, while the output is the SQL query. We adopt the pre-trained self-supervised speech representation model Hubert~\cite{hsu2021hubert} and language model GloVe~\cite{pennington2014glove} as the backbone of our model.
\paragraph{\textbf{Training.}}
\par The final loss terms in training are composed of the two parts:1) domain classification loss $\mathcal{L}_{CE}$: cross-entropy loss between the predicted speaker ID and the ground-truth; 2) SQL generation loss $\mathcal{L}_{MLE}$: maximum likelihood estimation(MLE) based on the given SQL query to maximize the predicted probability $p(y|x, a_{<t})$ based on a given SQL query. Note that, the domain classification loss $\mathcal{L}_{CE}$ is trained to remove speaker information but preserve semantic information, which is helpful for the final objective $\mathcal{L}_{MLE}$ to generate more accurate SQL query and the loss weight of $\mathcal{L}_{CE}$ is set to be 0.01.

\paragraph{\textbf{Inference.}}
\par After training, for each pair of the spoken question and database schema, we generate the target SQL query according to grammar rules with heuristics decode. We replace the special tokens in the target sequences with the SQL keywords.

\begin{table*}[ht]
    \centering
    \small
    \vspace{-3mm}
    \begin{tabular}{lcccccccc}
        \toprule
        \bfseries Method & SELECT & WHERE & GROUP & ORDER &  AND/OR & KEYWORD & Exact Match \\
        \midrule
        \bfseries Model Performance \\
        \midrule
        S2SQL-TTS     & 67.4 & 50.0 & 66.3 & 67.5 & 97.1 & 78.1 & 38.0     \\
        \midrule
        DeepSpeechSQL & 43.3 & 25.8 & 26.2 & 17.5 & 93.9 & 35.1 & 14.4   \\
        Cascaded    & 58.6 & 44.2 & \bfseries 56.4 & 59.4 & 96.2  & \bfseries 72.4 & 31.6     \\
         \bfseries Wav2SQL & \bfseries 63.6 & \bfseries  47.3 & 56.3 & \bfseries 60.2 & \bfseries 96.6 & 71.1 & \bfseries 34.1\\
        \midrule
        \bfseries Generalization to Custom Data \\
        \midrule
        S2SQL-TTS     & 67.8 & 45.5 & 53.6 & 69.2 & 96.6 & 73.1 & 35.7     \\
        \midrule
        DeepSpeechSQL & 41.0 & 24.2 & 25.5 & 18.1 & 95.1  & 31.8 & 12.2     \\
        Cascaded    & \bfseries 61.9 & \bfseries 45.2 &  50.5 & 61.7 & \bfseries 96.2  & \bfseries 73.3 & \bfseries 30.4     \\
         \bfseries Wav2SQL & 61.1 & 38.6 & \bfseries 51.5 & \bfseries 62.3 & 95.8 & 63.1 & 28.6 \\
        \midrule
        \bfseries Ablation Study \\
        \midrule
        w/o Speech reprogramming & 59.6 & 39.7 & 48.0 & 53.6 & 95.8 & 63.4    & 26.7  \\
        w/o Gradient reversal classifier & 62.1 & 37.9 & 52.6 & 60.7 & 95.8 & 63.2 & 27.5  \\
        
        \bottomrule
    \end{tabular}
    \vspace{-2mm}
    \caption{F1 scores of component matching and exact match accuracy on the MASpider test set comparison with baseline systems. We adopt Hubert as the speech feature extractor and GloVe as the language model.}
    \vspace{-3mm}
    \label{res:1}
\end{table*}

\section {Experiments}
\subsection{Experimental Setup}

\textbf{Dataset and Evaluation Metrics} MASpider consists of 9693 spoken questions from 11 speakers. Following the common practice~\cite{yu2018spider}, we evaluate the performance by exact match accuracy and component matching accuracy provided by~\cite{yu2018spider}, where exact match accuracy measures whether the predicted query is equivalent to the gold query as a whole while component matching measures the average exact match between the prediction and ground truth on different SQL components.

\textbf{Training and Inference}
We train our model on a single 32G NVIDIA V100 GPU with a batch size of 10 and a gradient accumulation value of 4 for 30k steps using the Adam optimizer with the default hyperparameters. The learning rate increases from 0 to $7.4 \times 10^{-4}$ with training. We preprocess column names and table names for tokenization and lemmatization using StandfordNLP toolkit~\cite{manning2014stanford}. For the pre-trained models, We adopt Hubert-Base~\footnote{\url{https://github.com/facebookresearch/fairseq/tree/main/examples/hubert}} to provide speech representation and GloVe~\cite{pennington2014glove} for the textual schema. In inference, we adopt beam search decoding with beam size 1.


\textbf{Model Configurations} 
In the SQL decoder, we set the rule embedding size as 128 and the node type embedding size as 64. A comprehensive table of hyperparameters is available in Appendix~\ref{sec:arch} in the supplementary materials.

\textbf{Baseline models} We compare generated SQL queries of our Wav2SQL with other systems, including:
(1) Cascaded: the cascaded model composed of automatic speech recognition(ASR) and text-to-SQL parsing model, which adopts the wav2vec 2.0~\cite{baevski2020wav2vec} and RATSQL~\cite{wang2019rat}.
(2) S2SQL-TTS: the S2SQL model trained on the dataset synthesized as our upper bound, where S2SQL means the model after removing the speech reprogramming, and adversarial learning in our Wav2SQL. The TTS model we adopt here is FastSpeech 2~\cite{ren2020fastspeech}.
(3) DeepSpeechSQL: This baseline employs the speech encoder of DeepSpeech~\cite{amodei2016deep} to replace Hubert as audio features, where other modules are consistent with the S2SQL model.

\subsection{Model Performance}
  
For in-domain evaluation, we prepare spoken questions with seen accents and speakers according to different SQL components, including SELECT, WHERE, GROUP, ORDER, AND/OR, and KEYWORD following~\cite{yu2018spider}. The results are compiled and presented in Table~\ref{res:1}, and we have the following observations: Wav2SQL surpasses the cascaded system across almost all SQL component matching and exact match accuracy on all SQL queries. Specifically, the SELECT and WHERE component have increased significantly by 5.0\% and 3.1\% respectively, and exact match accuracy has increased by 2.5\%, demonstrating the effectiveness of our direct speech-to-SQL parsing model. It indicates that our direct S2SQL model avoids error compounding across subsystems. Besides, Wav2SQL greatly surpasses DeepSpeechSQL which proves the superiority of Hubert. Compare to the upper bound less variance dataset constructed by a single-speaker single-accent TTS system, we still achieve competitive performance, indicating the efficiency of our proposed techniques for reducing acoustic attributes and promoting generalization. 

\begin{table}[ht]
    \centering
    \small    
    \vspace{1mm}
    \begin{tabular}{l|c|c|c|c}
        \toprule
        Dataset     & Easy & Medium & Hard & Extra \\
        \midrule
        DeepSpeechSQL & 29.4 & 10.3 & 11.5 & 6.0 \\
        Cascaded    & 51.2 & 28.3 & \bfseries 26.4 & 16.9 \\
        Wav2SQL       & \bfseries 54.8 & \bfseries 29.1  &   25.3  &  \bfseries 25.3 \\
        \bottomrule
    \end{tabular}
    \caption{A comparison to the cascaded model and DeepSpeechSQL model in MASpider according to the level of difficulty.}
    \label{res:diff}
\end{table}

To further verify the effectiveness of our methods, we compare our model with the cascaded system and DeepSpeechSQL model. We group the parsing difficulty into easy, medium, hard, and extra according to the number of component selections and conditions of the target SQL queries. As illustrated in Table~\ref{res:diff}, we have the following observations: 

1) As parsing difficulty increases, a distinct degradation could be witnessed in generation accuracy; and 2) our direct speech-to-SQL parsing model outperforms the cascade baseline since it avoids error compounding across subsystems, demonstrating a large margin improvement by 8.4\%, especially in the extra hard part; and 3) Our model is far superior to DeepSpeechSQL at all levels, indicating that the self-supervised Hubert model is more suitable for our task than the traditional ASR Encoder.

\begin{table*}[h]
    \centering
    \small
    \begin{tabular}{l}
    \toprule
    \textbf{Medium}: Show name, country, age for all singers ordered by age from the oldest to the youngest. \\
    \textbf{Cascaded}: SELECT \textcolor{red}{singer.Country, singer.Age} FROM singer ORDER BY singer.Age Desc \\
    \textbf{Wav2SQL}: SELECT \textcolor{blue}{singer.Name, singer.Country, singer.Age} FROM singer ORDER BY singer.Age Desc \\
    \textbf{Gold SQL}: SELECT Name, Country, Age FROM singer ORDER BY Age Desc \\
    \midrule
    \textbf{Hard}: List all song names by singers above the average age.\\
    \textbf{Cascaded}: SELECT singer.Song\_Name FROM singer WHERE singer.Age < \textcolor{red}{'terminal' ORDER BY singer.Song\_Name Asc} \\
    \textbf{Wav2SQL}: SELECT singer.Song\_Name FROM singer WHERE singer.Age > \textcolor{blue}{(SELECT Avg(singer.Age) FROM singer} \\
    \textbf{Gold SQL}: SELECT Song\_Name FROM singer WHERE Age > (SELECT avg(Age) FROM singer) \\
    \midrule
    \textbf{Extra}: Find the average age of students who do not have any pet. \\
    \textbf{Cascaded}: SELECT \textcolor{red}{Student.Fname} FROM Student WHERE Student.StuID NOT IN (SELECT Has\_Pet.StuID FROM Has\_Pet)\\
    \textbf{Wav2SQL}: SELECT \textcolor{blue}{Avg(Student.Age)} FROM Student WHERE Student.StuID NOT IN (SELECT Has\_Pet.StuID FROM Has\_Pet) \\
    \textbf{Gold SQL}: SELECT avg(Age) from Student where stuid not in (select stuid from has\_pet) \\
    \bottomrule
    \end{tabular}
    \vspace{2mm}
    \caption{Three examples compared with cascaded system. We mark the wrong part of the cascaded model in red while the corresponding correct part in Wav2SQL in blue. The input question is represented by SQL query difficulty. Cell values in the SQL queries are replaced with placeholder "terminal".}
    \label{table:case}
\vspace{-3mm}
    \end{table*}
    
\subsection{Generalization To Custom Data}
For out-of-domain testing, we prepare spoken questions with databases, accents, and speakers that are unseen in custom data. The results are summarized in Table~\ref{res:1}, and we have the following observations:  1) As shown in the table, we see that our proposed Wav2SQL outperforms DeepSpeechSQL by a large margin of 16.4\% on exact match accuracy. In addition, the component matching of our model on all SQL components outperforms DeepSpeechSQL, especially in SELECT, GROUP BY, and ORDER BY components by 20.1\%, 26.0\%, and 44.2\%; 2) 
Under the challenge of invisible accents, Wav2SQL can still maintain competitive results with a 1.8\% exact match accuracy drop compared with the cascaded system, which validates the superiority of our model by exploiting speech re-program and adversarial training to get deterministic representations invariant to accents and speakers; 3) Although we are pleasantly surprised to find that we have a small gap with S2SQL-TTS in ELECT, GROUP BY and AND/OR component matching, there is still a certain gap with it in exact match accuracy, and we will further narrow this gap in future work.

\subsection{Ablation Studies}
As shown in Table~\ref{res:1}, we conduct ablation studies to demonstrate the effectiveness of several designs in our model, including speech re-programming and gradient reversal classifier technique, and text schema linking. The results have been presented in Table \ref{res:1}, and we have the following discovering: 1) the speech re-programming methods show an improvement in exact match accuracy by 1.9\% and a substantial increase of 8.7\% in ORDER BY component matching, indicating its efficiency in reducing acoustic variance and learning deterministic representations; 2) Removing the gradient reversal classifier has witnessed a distinct degradation in model performance by 1.1\% accuracy especially in ORDER By component matching (1.6\%), showing its superiority in learning speaker-agnostic speech representation; 

\subsection{Case Study}

We compare the SQL query generated by Wav2SQL with the baseline cascaded system in Table~\ref{table:case}. As shown in the table, we can notice that Wav2SQL has better performance than the baseline system, mainly in SELECT, WHERE, and KEYWORDS operations. For example, in the first and third cases, the cascading system fails to fill the values into the correct slots, thus, it stupidly forgets the 'Name' of table 'Singer' and is unable to select the correct column name(i.e, "Age"). In addition, Wav2SQL successfully completes the averaging operation on "Age" in the third case. Unfortunately, in the second example, the cascaded system incorrectly constructs the WHERE clause so that it fails to pick singers who are older than the average age.

\section{Conclusion}
We released MASpider, the first large-scale, multi-speaker, and multi-accent S2SQL parsing dataset, which we hope would accelerate S2SQL research in the community. In this work, we presented the first direct speech-to-SQL model Wav2SQL which avoided error compounding across cascaded systems. To tackle the data scarcity issue, we leveraged recent progress in large-scale pre-training and utilized self-supervised models to derive discriminate representation. To promote model generalization and robustness to custom out-of-distribution data, we further introduced speech re-programming and gradient-reversal classifier techniques which reduced acoustic variance and learned style-agnostic representations. Experimental results demonstrated that our approach achieved new state-of-the-art results by up to 4.6\% accuracy improvement over baseline. In the future, we will investigate techniques to further enhance the model generalization in direct Speech-to-SQL parsing.

\section{Limitation and Potential Risks}
As mentioned in the model performance, there is still a certain gap between Wav2SQL and S2SQL-TTS. One of our future directions is to further remove accent and speaker information to improve generation performance. In addition, our experiments find that the schema linking we adopt is still rough compared to text schema linking, which seriously affects the performance of our model. In the future work, we will study how to obtain accurate and fine-grained schema linking.

Wav2SQL lowers the requirements for speech-to-SQL generation, which may cause unemployment for people with related occupations databases developers and SQL programmer. Furthermore, there is the potential for leading to the misuse of database than they expect.

\bibliography{anthology,custom}

\begin{thebibliography}{46}
\expandafter\ifx\csname natexlab\endcsname\relax\def\natexlab#1{#1}\fi

\bibitem[{Amodei et~al.(2016)Amodei, Ananthanarayanan, Anubhai, Bai,
  Battenberg, Case, Casper, Catanzaro, Cheng, Chen et~al.}]{amodei2016deep}
Dario Amodei, Sundaram Ananthanarayanan, Rishita Anubhai, Jingliang Bai, Eric
  Battenberg, Carl Case, Jared Casper, Bryan Catanzaro, Qiang Cheng, Guoliang
  Chen, et~al. 2016.
\newblock Deep speech 2: End-to-end speech recognition in english and mandarin.
\newblock In \emph{International conference on machine learning}, pages
  173--182. PMLR.

\bibitem[{Baevski et~al.(2019)Baevski, Schneider, and Auli}]{baevski2019vq}
Alexei Baevski, Steffen Schneider, and Michael Auli. 2019.
\newblock vq-wav2vec: Self-supervised learning of discrete speech
  representations.
\newblock \emph{arXiv preprint arXiv:1910.05453}.

\bibitem[{Baevski et~al.(2020)Baevski, Zhou, Mohamed, and
  Auli}]{baevski2020wav2vec}
Alexei Baevski, Yuhao Zhou, Abdelrahman Mohamed, and Michael Auli. 2020.
\newblock wav2vec 2.0: A framework for self-supervised learning of speech
  representations.
\newblock \emph{Advances in Neural Information Processing Systems},
  33:12449--12460.

\bibitem[{Bogin et~al.(2019{\natexlab{a}})Bogin, Gardner, and
  Berant}]{bogin2019global}
Ben Bogin, Matt Gardner, and Jonathan Berant. 2019{\natexlab{a}}.
\newblock Global reasoning over database structures for text-to-sql parsing.
\newblock \emph{arXiv preprint arXiv:1908.11214}.

\bibitem[{Bogin et~al.(2019{\natexlab{b}})Bogin, Gardner, and
  Berant}]{bogin2019representing}
Ben Bogin, Matt Gardner, and Jonathan Berant. 2019{\natexlab{b}}.
\newblock Representing schema structure with graph neural networks for
  text-to-sql parsing.
\newblock \emph{arXiv preprint arXiv:1905.06241}.

\bibitem[{Bui et~al.(2021)Bui, Tran, Tran, and Phung}]{bui2021exploiting}
Manh-Ha Bui, Toan Tran, Anh Tran, and Dinh Phung. 2021.
\newblock Exploiting domain-specific features to enhance domain generalization.
\newblock \emph{Advances in Neural Information Processing Systems},
  34:21189--21201.

\bibitem[{Cai et~al.(2021)Cai, Yuan, Xu, and Hao}]{cai2021sadga}
Ruichu Cai, Jinjie Yuan, Boyan Xu, and Zhifeng Hao. 2021.
\newblock Sadga: Structure-aware dual graph aggregation network for
  text-to-sql.
\newblock \emph{Advances in Neural Information Processing Systems},
  34:7664--7676.

\bibitem[{Campbell(2008)}]{campbell2008ethnologue}
Lyle Campbell. 2008.
\newblock Ethnologue: Languages of the world.

\bibitem[{Chen et~al.(2020)Chen, Lin, Li, Li, Zhou, and
  Sun}]{chen2020measuring}
Deli Chen, Yankai Lin, Wei Li, Peng Li, Jie Zhou, and Xu~Sun. 2020.
\newblock Measuring and relieving the over-smoothing problem for graph neural
  networks from the topological view.
\newblock In \emph{Proceedings of the AAAI Conference on Artificial
  Intelligence}, volume~34, pages 3438--3445.

\bibitem[{Choi et~al.(2021)Choi, Lee, Kim, Lee, Heo, and Lee}]{choi2021neural}
Hyeong-Seok Choi, Juheon Lee, Wansoo Kim, Jie Lee, Hoon Heo, and Kyogu Lee.
  2021.
\newblock Neural analysis and synthesis: Reconstructing speech from
  self-supervised representations.
\newblock \emph{Advances in Neural Information Processing Systems},
  34:16251--16265.

\bibitem[{Ganin et~al.(2016)Ganin, Ustinova, Ajakan, Germain, Larochelle,
  Laviolette, Marchand, and Lempitsky}]{ganin2016domain}
Yaroslav Ganin, Evgeniya Ustinova, Hana Ajakan, Pascal Germain, Hugo
  Larochelle, Fran{\c{c}}ois Laviolette, Mario Marchand, and Victor Lempitsky.
  2016.
\newblock Domain-adversarial training of neural networks.
\newblock \emph{The journal of machine learning research}, 17(1):2096--2030.

\bibitem[{Guo et~al.(2019)Guo, Zhan, Gao, Xiao, Lou, Liu, and
  Zhang}]{guo2019towards}
Jiaqi Guo, Zecheng Zhan, Yan Gao, Yan Xiao, Jian-Guang Lou, Ting Liu, and
  Dongmei Zhang. 2019.
\newblock Towards complex text-to-sql in cross-domain database with
  intermediate representation.
\newblock \emph{arXiv preprint arXiv:1905.08205}.

\bibitem[{Hsu et~al.(2021)Hsu, Bolte, Tsai, Lakhotia, Salakhutdinov, and
  Mohamed}]{hsu2021hubert}
Wei-Ning Hsu, Benjamin Bolte, Yao-Hung~Hubert Tsai, Kushal Lakhotia, Ruslan
  Salakhutdinov, and Abdelrahman Mohamed. 2021.
\newblock Hubert: Self-supervised speech representation learning by masked
  prediction of hidden units.
\newblock \emph{IEEE/ACM Transactions on Audio, Speech, and Language
  Processing}, 29:3451--3460.

\bibitem[{Huang et~al.(2022{\natexlab{a}})Huang, Cui, Chen, Ren, Liu, Zhao,
  Huai, and Wang}]{huang2022singgan}
Rongjie Huang, Chenye Cui, Feiyang Chen, Yi~Ren, Jinglin Liu, Zhou Zhao,
  Baoxing Huai, and Zhefeng Wang. 2022{\natexlab{a}}.
\newblock Singgan: Generative adversarial network for high-fidelity singing
  voice generation.
\newblock In \emph{Proceedings of the 30th ACM International Conference on
  Multimedia}, pages 2525--2535.

\bibitem[{Huang et~al.(2023{\natexlab{a}})Huang, Huang, Yang, Ren, Liu, Li, Ye,
  Liu, Yin, and Zhao}]{huang2023make}
Rongjie Huang, Jiawei Huang, Dongchao Yang, Yi~Ren, Luping Liu, Mingze Li,
  Zhenhui Ye, Jinglin Liu, Xiang Yin, and Zhou Zhao. 2023{\natexlab{a}}.
\newblock Make-an-audio: Text-to-audio generation with prompt-enhanced
  diffusion models.
\newblock \emph{arXiv preprint arXiv:2301.12661}.

\bibitem[{Huang et~al.(2022{\natexlab{b}})Huang, Lam, Wang, Su, Yu, Ren, and
  Zhao}]{huang2022fastdiff}
Rongjie Huang, Max~WY Lam, Jun Wang, Dan Su, Dong Yu, Yi~Ren, and Zhou Zhao.
  2022{\natexlab{b}}.
\newblock Fastdiff: A fast conditional diffusion model for high-quality speech
  synthesis.
\newblock \emph{arXiv preprint arXiv:2204.09934}.

\bibitem[{Huang et~al.(2023{\natexlab{b}})Huang, Li, Yang, Shi, Chang, Ye, Wu,
  Hong, Huang, Liu et~al.}]{huang2023audiogpt}
Rongjie Huang, Mingze Li, Dongchao Yang, Jiatong Shi, Xuankai Chang, Zhenhui
  Ye, Yuning Wu, Zhiqing Hong, Jiawei Huang, Jinglin Liu, et~al.
  2023{\natexlab{b}}.
\newblock Audiogpt: Understanding and generating speech, music, sound, and
  talking head.
\newblock \emph{arXiv preprint arXiv:2304.12995}.

\bibitem[{Huang et~al.(2022{\natexlab{c}})Huang, Ren, Liu, Cui, and
  Zhao}]{huang2022generspeech}
Rongjie Huang, Yi~Ren, Jinglin Liu, Chenye Cui, and Zhou Zhao.
  2022{\natexlab{c}}.
\newblock Generspeech: Towards style transfer for generalizable out-of-domain
  text-to-speech synthesis.
\newblock \emph{arXiv preprint arXiv:2205.07211}.

\bibitem[{Huang et~al.(2022{\natexlab{d}})Huang, Zhao, Liu, Liu, Cui, and
  Ren}]{huang2022prodiff}
Rongjie Huang, Zhou Zhao, Huadai Liu, Jinglin Liu, Chenye Cui, and Yi~Ren.
  2022{\natexlab{d}}.
\newblock Prodiff: Progressive fast diffusion model for high-quality
  text-to-speech.
\newblock \emph{arXiv preprint arXiv:2207.06389}.

\bibitem[{Huang et~al.(2022{\natexlab{e}})Huang, Zhao, Liu, Liu, Ren, Zhang,
  and He}]{huang2022transpeech}
Rongjie Huang, Zhou Zhao, Jinglin Liu, Huadai Liu, Yi~Ren, Lichao Zhang, and
  Jinzheng He. 2022{\natexlab{e}}.
\newblock Transpeech: Speech-to-speech translation with bilateral perturbation.
\newblock \emph{arXiv preprint arXiv:2205.12523}.

\bibitem[{Jia et~al.(2019)Jia, Weiss, Biadsy, Macherey, Johnson, Chen, and
  Wu}]{jia2019direct}
Ye~Jia, Ron~J Weiss, Fadi Biadsy, Wolfgang Macherey, Melvin Johnson, Zhifeng
  Chen, and Yonghui Wu. 2019.
\newblock Direct speech-to-speech translation with a sequence-to-sequence
  model.
\newblock \emph{arXiv preprint arXiv:1904.06037}.

\bibitem[{Kumar et~al.(2013)Kumar, Kumar, Mitra, and
  Sundaram}]{kumar2013system}
Sachin Kumar, Ashish Kumar, Pinaki Mitra, and Girish Sundaram. 2013.
\newblock System and methods for converting speech to sql.
\newblock \emph{arXiv preprint arXiv:1308.3106}.

\bibitem[{Lam et~al.(2021)Lam, Wang, Huang, Su, and Yu}]{lam2021bilateral}
Max~WY Lam, Jun Wang, Rongjie Huang, Dan Su, and Dong Yu. 2021.
\newblock Bilateral denoising diffusion models.
\newblock \emph{arXiv preprint arXiv:2108.11514}.

\bibitem[{Lee et~al.(2021)Lee, Chen, Wang, Gu, Ma, Polyak, Adi, He, Tang, Pino
  et~al.}]{lee2021direct}
Ann Lee, Peng-Jen Chen, Changhan Wang, Jiatao Gu, Xutai Ma, Adam Polyak, Yossi
  Adi, Qing He, Yun Tang, Juan Pino, et~al. 2021.
\newblock Direct speech-to-speech translation with discrete units.
\newblock \emph{arXiv preprint arXiv:2107.05604}.

\bibitem[{Li et~al.(2018{\natexlab{a}})Li, Pan, Wang, and Kot}]{li2018domain}
Haoliang Li, Sinno~Jialin Pan, Shiqi Wang, and Alex~C Kot. 2018{\natexlab{a}}.
\newblock Domain generalization with adversarial feature learning.
\newblock In \emph{Proceedings of the IEEE conference on computer vision and
  pattern recognition}, pages 5400--5409.

\bibitem[{Li et~al.(2018{\natexlab{b}})Li, Tian, Gong, Liu, Liu, Zhang, and
  Tao}]{li2018deep}
Ya~Li, Xinmei Tian, Mingming Gong, Yajing Liu, Tongliang Liu, Kun Zhang, and
  Dacheng Tao. 2018{\natexlab{b}}.
\newblock Deep domain generalization via conditional invariant adversarial
  networks.
\newblock In \emph{Proceedings of the European Conference on Computer Vision
  (ECCV)}, pages 624--639.

\bibitem[{Liu et~al.(2021)Liu, Li, and Lee}]{liu2021tera}
Andy~T Liu, Shang-Wen Li, and Hung-yi Lee. 2021.
\newblock Tera: Self-supervised learning of transformer encoder representation
  for speech.
\newblock \emph{IEEE/ACM Transactions on Audio, Speech, and Language
  Processing}, 29:2351--2366.

\bibitem[{Manning et~al.(2014)Manning, Surdeanu, Bauer, Finkel, Bethard, and
  McClosky}]{manning2014stanford}
Christopher~D Manning, Mihai Surdeanu, John Bauer, Jenny~Rose Finkel, Steven
  Bethard, and David McClosky. 2014.
\newblock The stanford corenlp natural language processing toolkit.
\newblock In \emph{Proceedings of 52nd annual meeting of the association for
  computational linguistics: system demonstrations}, pages 55--60.

\bibitem[{Nakamura et~al.(2006)Nakamura, Markov, Nakaiwa, Kikui, Kawai,
  Jitsuhiro, Zhang, Yamamoto, Sumita, and Yamamoto}]{nakamura2006atr}
Satoshi Nakamura, Konstantin Markov, Hiromi Nakaiwa, Gen-ichiro Kikui, Hisashi
  Kawai, Takatoshi Jitsuhiro, J-S Zhang, Hirofumi Yamamoto, Eiichiro Sumita,
  and Seiichi Yamamoto. 2006.
\newblock The atr multilingual speech-to-speech translation system.
\newblock \emph{IEEE Transactions on Audio, Speech, and Language Processing},
  14(2):365--376.

\bibitem[{Panayotov et~al.(2015)Panayotov, Chen, Povey, and
  Khudanpur}]{panayotov2015librispeech}
Vassil Panayotov, Guoguo Chen, Daniel Povey, and Sanjeev Khudanpur. 2015.
\newblock Librispeech: an asr corpus based on public domain audio books.
\newblock In \emph{2015 IEEE international conference on acoustics, speech and
  signal processing (ICASSP)}, pages 5206--5210. IEEE.

\bibitem[{Pennington et~al.(2014)Pennington, Socher, and
  Manning}]{pennington2014glove}
Jeffrey Pennington, Richard Socher, and Christopher~D Manning. 2014.
\newblock Glove: Global vectors for word representation.
\newblock In \emph{Proceedings of the 2014 conference on empirical methods in
  natural language processing (EMNLP)}, pages 1532--1543.

\bibitem[{Polyak and Wolf(2019)}]{polyak2019attention}
Adam Polyak and Lior Wolf. 2019.
\newblock Attention-based wavenet autoencoder for universal voice conversion.
\newblock In \emph{ICASSP 2019-2019 IEEE International Conference on Acoustics,
  Speech and Signal Processing (ICASSP)}, pages 6800--6804. IEEE.

\bibitem[{Qian et~al.(2020)Qian, Zhang, Chang, Hasegawa-Johnson, and
  Cox}]{qian2020unsupervised}
Kaizhi Qian, Yang Zhang, Shiyu Chang, Mark Hasegawa-Johnson, and David Cox.
  2020.
\newblock Unsupervised speech decomposition via triple information bottleneck.
\newblock In \emph{International Conference on Machine Learning}, pages
  7836--7846. PMLR.

\bibitem[{Ren et~al.(2020)Ren, Hu, Tan, Qin, Zhao, Zhao, and
  Liu}]{ren2020fastspeech}
Yi~Ren, Chenxu Hu, Xu~Tan, Tao Qin, Sheng Zhao, Zhou Zhao, and Tie-Yan Liu.
  2020.
\newblock Fastspeech 2: Fast and high-quality end-to-end text to speech.
\newblock \emph{arXiv preprint arXiv:2006.04558}.

\bibitem[{Schneider et~al.(2019)Schneider, Baevski, Collobert, and
  Auli}]{schneider2019wav2vec}
Steffen Schneider, Alexei Baevski, Ronan Collobert, and Michael Auli. 2019.
\newblock wav2vec: Unsupervised pre-training for speech recognition.
\newblock \emph{arXiv preprint arXiv:1904.05862}.

\bibitem[{Shi et~al.(2021)Shi, Seely, Torr, Siddharth, Hannun, Usunier, and
  Synnaeve}]{shi2021gradient}
Yuge Shi, Jeffrey Seely, Philip~HS Torr, N~Siddharth, Awni Hannun, Nicolas
  Usunier, and Gabriel Synnaeve. 2021.
\newblock Gradient matching for domain generalization.
\newblock \emph{arXiv preprint arXiv:2104.09937}.

\bibitem[{Song et~al.(2022)Song, Wong, Zhao, and Jiang}]{song2022speech}
Yuanfeng Song, Raymond Chi-Wing Wong, Xuefang Zhao, and Di~Jiang. 2022.
\newblock Speech-to-sql: Towards speech-driven sql query generation from
  natural language question.
\newblock \emph{arXiv preprint arXiv:2201.01209}.

\bibitem[{Tian et~al.(2022)Tian, Li, Xie, Liu, and Wang}]{tian2022neuron}
Chris~Xing Tian, Haoliang Li, Xiaofei Xie, Yang Liu, and Shiqi Wang. 2022.
\newblock Neuron coverage-guided domain generalization.
\newblock \emph{IEEE Transactions on Pattern Analysis and Machine
  Intelligence}.

\bibitem[{Vaswani et~al.(2017)Vaswani, Shazeer, Parmar, Uszkoreit, Jones,
  Gomez, Kaiser, and Polosukhin}]{vaswani2017attention}
Ashish Vaswani, Noam Shazeer, Niki Parmar, Jakob Uszkoreit, Llion Jones,
  Aidan~N Gomez, {\L}ukasz Kaiser, and Illia Polosukhin. 2017.
\newblock Attention is all you need.
\newblock \emph{Advances in neural information processing systems}, 30.

\bibitem[{Wang et~al.(2019)Wang, Shin, Liu, Polozov, and
  Richardson}]{wang2019rat}
Bailin Wang, Richard Shin, Xiaodong Liu, Oleksandr Polozov, and Matthew
  Richardson. 2019.
\newblock Rat-sql: Relation-aware schema encoding and linking for text-to-sql
  parsers.
\newblock \emph{arXiv preprint arXiv:1911.04942}.

\bibitem[{Yin and Neubig(2017)}]{yin2017syntactic}
Pengcheng Yin and Graham Neubig. 2017.
\newblock A syntactic neural model for general-purpose code generation.
\newblock \emph{arXiv preprint arXiv:1704.01696}.

\bibitem[{Yu and Deng(2016)}]{yu2016automatic}
Dong Yu and Li~Deng. 2016.
\newblock \emph{Automatic speech recognition}, volume~1.
\newblock Springer.

\bibitem[{Yu et~al.(2018)Yu, Zhang, Yang, Yasunaga, Wang, Li, Ma, Li, Yao,
  Roman et~al.}]{yu2018spider}
Tao Yu, Rui Zhang, Kai Yang, Michihiro Yasunaga, Dongxu Wang, Zifan Li, James
  Ma, Irene Li, Qingning Yao, Shanelle Roman, et~al. 2018.
\newblock Spider: A large-scale human-labeled dataset for complex and
  cross-domain semantic parsing and text-to-sql task.
\newblock \emph{arXiv preprint arXiv:1809.08887}.

\bibitem[{Zhao et~al.(2020)Zhao, Gong, Liu, Fu, and Tao}]{zhao2020domain}
Shanshan Zhao, Mingming Gong, Tongliang Liu, Huan Fu, and Dacheng Tao. 2020.
\newblock Domain generalization via entropy regularization.
\newblock \emph{Advances in Neural Information Processing Systems},
  33:16096--16107.

\bibitem[{Zhong et~al.(2017)Zhong, Xiong, and Socher}]{zhong2017seq2sql}
Victor Zhong, Caiming Xiong, and Richard Socher. 2017.
\newblock Seq2sql: Generating structured queries from natural language using
  reinforcement learning.
\newblock \emph{arXiv preprint arXiv:1709.00103}.

\bibitem[{Zhou et~al.(2020)Zhou, Yang, Hospedales, and Xiang}]{zhou2020deep}
Kaiyang Zhou, Yongxin Yang, Timothy Hospedales, and Tao Xiang. 2020.
\newblock Deep domain-adversarial image generation for domain generalisation.
\newblock In \emph{Proceedings of the AAAI Conference on Artificial
  Intelligence}, volume~34, pages 13025--13032.

\end{thebibliography}

\appendix
\clearpage

\begin{center}
    {\bf {\LARGE Appendices}}
\end{center}

\section{Domain Classifier}
\label{sec:domain}

Domain classifier effectively captures the audio's long-term speaker identity and predicts the speaker ID for the spoken question. After training on augmented data, the domain classifier could attain robust representations that capture an ample speaker identity space. Combined with gradient reversal, we can get deterministic representation agnostic to speaker discrepancy, significantly reducing intro-speaker variance and making it possible for tree-structured depth-first decoding.

\section{Acoustic Perturbation}\label{sec:ap}
To obtain speech samples with acoustic information enhancement, we adopt the following functions~\cite{huang2022transpeech,qian2020unsupervised, choi2021neural} to perturb the acoustic features, that is 1) random resampling $RR$, and 2) formant shifting $fs$, and 3)pitch randomization $pr$, 4) random frequency shaping using a parametric equalizer $peq$. 
\begin{itemize}[leftmargin=*]
    \item For $RR$, a random resampling is adopted to modify the rhythm. The raw waveform is divided into segments, whose length is randomly uniformly drawn from 19 frames to 32 frames~\cite{polyak2019attention}. Each segment is resampled using linear interpolation with a resampling factor randomly drawn from 0.5 to 1.5.
    \item For $fs$, a formant shifting ratio is sampled uniformly from $\mathrm{Uniform}(1,1.4)$. After sampling the ratio, we again randomly decided whether to take the reciprocal of the sampled ratio or not. 
    \item For $pr$, a pitch shift ratio and a pitch range ratio are sampled uniformly from $\mathrm{Uniform}(1,2)$ and $\mathrm{Uniform}(1,1.5)$, respectively. Again, we randomly decide whether to take the reciprocal of the sampled ratios or not. For more details on formant shifting and pitch randomization, please refer to Parselmouth~\url{https://github.com/YannickJadoul/Parselmouth}.
    \item Lastly, $peq$ denotes a serial composition of low-shelving, peaking, and high-shelving filters. We use one low-shelving HLS, one high-shelving HHS, and eight peaking filters HPeak.
\end{itemize}

\section{Model Architectures}
\label{sec:arch}
\begin{figure*}[ht]
    \small
    \centering
    \includegraphics[scale=0.6]{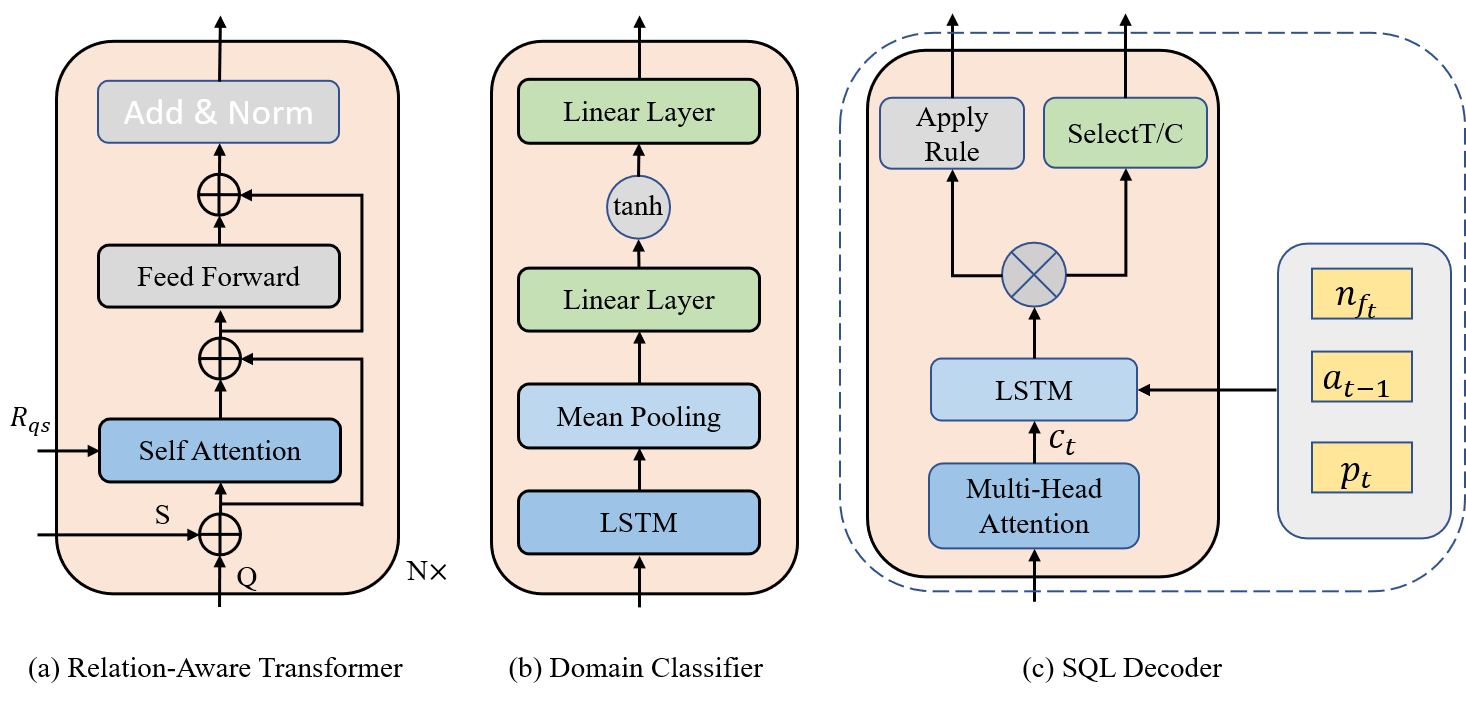}
    \caption{Architecture for relation-aware transformer, SQL decoder, and domain classifier. The self-attention hear is relation-aware. Q: spoken question; S: schema; $R_ij$: relations from any question item or schema item; SelectT/C: SELECTTABLE/SELECTCOLUMN.}
    \label{fig:app}
\end{figure*}

We list the model hyperparameters of Wav2SQL in Table~\ref{tab:hyperparameters} and illustrate the architecture for relational-aware transformer(RAT), SQL decoder, and domain classifier in Figure~\ref{fig:app}. The schema linking used by RAT in the train set is borrowed from RATSQL\cite{wang2019rat} while the schema linking of test set comes from string matching between the text obtained by ASR and the schema.

\textbf{Hubert}
The Hubert feature extractor consists of seven blocks and the temporal convolutions in each block have 512 channels with strides (5,2,2,2,2,2,2) and kernel widths (10,3,3,3,3,2,2), and 12 transformer blocks, model dimension 768, inner dimension (FFN) 3,072 and 8 attention heads.

\textbf{Relation-Aware Transformer}
The relation-aware encoder consists of 8 transformer layers. Each layer contains a relation-aware self-attention module, the final output passes through a feed-forward layer. 

\begin{table}[htbp]
    \small
    \centering
    \begin{tabular}{l|c|c}
        \toprule
        \multicolumn{2}{c|}{Hyperparameter}                       & Wav2SQL      \\
        \midrule
        \multirow{3}{*}{Speech Encoder}
                                            & Hubert Hidden     & 512        \\
                                            & LSTM Hidden      & 256        \\
                                            & LSTM Layers      & 3          \\
        \midrule
        \multirow{3}{*}{Text Encoder}
                                            & GloVe Embedding    & 512        \\
                                            & LSTM Hidden      & 256        \\
                                            & LSTM Layers      & 1          \\
        \midrule
        \multirow{2}{*}{Domain Classifier}
                                            & LSTM Hidden    & 256 \\
                                            & LSTM Layers         & 2    \\
                                            & LSTM Dropout        & 0.2   \\
                                            & Output Dimension    & 11   \\
        \midrule
        \multirow{5}{*}{Relation-Aware Transformer}
                                            & Transformer Block               & 8     \\
                                            & Hidden  Size            & 256    \\
                                            & Attention Heads     & 8       \\
                                            & FFN Size           & 1024 \\
                                            & Dropout             & 0.1    \\
        \midrule
        \multirow{4}{*}{SQL Decoder}
                                            & Rule Embedding    & 128   \\
                                            & Node Embedding    & 64      \\
                                            & LSTM Size           & 512   \\
                                            & Dropout             & 0.21  \\
        \bottomrule
    \end{tabular}
    \vspace{2mm}
    \caption{Hyperparameters of Wav2SQL.}
    \label{tab:hyperparameters}
\end{table}

\section{Ethical Considerations}
Our MASpider benchmark presented in this work is a free and open source for the community to study speech-to-SQL parsing. We collect and annotate recordings from the mainstream text-to-SQL dataset, Spider~\cite{yu2018spider}, which is also free and open datasets for research use. For audio recording, we hire annotators from different countries to record audio in a quiet environment. We pay the annotators an average of 80 dollars per hour.

\end{document}